\documentclass[runningheads]{llncs}

\usepackage{eccv}

\usepackage{eccvabbrv}

\usepackage{graphicx}
\usepackage{booktabs}
\usepackage{multirow}
\usepackage{amsmath}
\usepackage{makecell}

\usepackage[accsupp]{axessibility}  

\usepackage{orcidlink}
\usepackage{siunitx}
\DeclareSIUnit{\ppm}{ppm}
\usepackage{tabularx}

\begin{document}

\title{Uncertainty-aware tree height change regression} 


\author{Max Gaber\inst{1}\orcidlink{0009-0002-7116-1496} \and
Dimitri Gominski\inst{1}\orcidlink{0000-0002-8135-1341} \and
Jaime C. Revenga\inst{2}\orcidlink{0000-0002-9330-6572} \and
Stefan Oehmcke\inst{3,4}\orcidlink{0000-0002-0240-1559} \and
Rasmus Fensholt\inst{1}\orcidlink{0000-0003-3067-4527} \and
Martin Brandt\inst{1}\orcidlink{0000-0001-9531-1239}}

\authorrunning{M.~Gaber et al.}

\institute{Department of Geosciences and Natural Resource Management, University of Copenhagen\and
Department of Mathematical Sciences, University of Copenhagen\and
Department of Computer Science and Electrical Engineering, University of Rostock\and
Department of Computer Science, University of Copenhagen}

\maketitle

\begin{abstract}
Monitoring canopy height change is essential for understanding carbon sinks and forest dynamics. Remote sensing enables consistent, large-scale observations of such changes, increasingly integrated with deep learning architectures such as Geospatial Foundation Models (GFMs). However, existing methods and datasets frame the problem as binary change detection, which overlooks both the continuous nature of change, especially for vegetation, and the inherent uncertainty in labels. We present the \textit{Canopy Height Change (CHC)} dataset, providing 3 m resolution \textit{continuous} canopy height differences and associated spatially resolved uncertainties across \SI{10598}{\square\kilo\meter} of northern and western Spain. The dataset is paired with a co-located time series of PlanetScope satellite imagery.  Based on the dataset, we introduce the task of uncertainty-aware change regression, associated metrics and strategies for fine-tuning GFMs. Furthermore, we evaluate state-of-the-art GFMs and highlight promising directions and remaining challenges for advancing continuous canopy height change estimation.
\end{abstract}

\section{Introduction}
\label{sec:intro}
Understanding how canopy height changes over time is essential for quantifying carbon sinks and sources, ecosystem dynamics, monitoring disturbances, and detecting tree growth or deforestation. Recent advances in remote sensing, particularly in high resolution satellite imagery and deep learning, now enable consistent characterization of canopy height as wall-to-wall maps at continental and global scales \cite{lang_high-resolution_2023, liu_overlooked_2023, pauls_capturing_2025, tolan_sub-meter_2023}. Typically, these maps are produced through computer vision models, trained in a supervised approach on satellite imagery and target data originating from Airborne Laser Scanning (ALS) or spaceborne LiDAR, such as GEDI, over a wide range of geographies and biomes.\par
\begin{figure}
    \centering
    \includegraphics[width=0.9\linewidth]{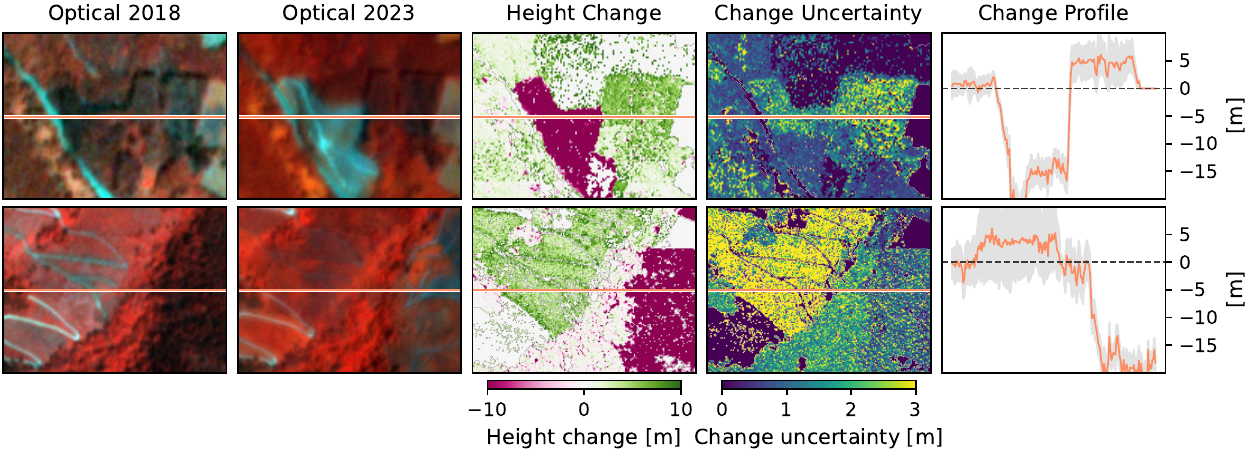}
    \caption{The Canopy Height Change (CHC) dataset consists of a time series of optical PlanetScope images (6 years of min. 1 image, displayed here as false color composite), co-located with continuous pixel-wise canopy height change data and a height change uncertainty layer. It enables benchmarking of continuous canopy height change predictions while explicitly accounting for uncertainty in the reference data. The figure shows the change profile along the red transectional line with \SI{90}{\percent} confidence interval for illustrative purposes.}
    \label{fig:overview}
\end{figure}
Recently, increasing attention has been directed toward estimating the temporal dynamics of canopy height, leading to the development of geospatial models that incorporate a time dimension \cite{pauls_capturing_2025, pauls_echosat_2026, schwartz_retrieving_2025}. However, the validation data of the resulting products is not public, and either limited to single-time-step evaluation or change evaluation without quantification of uncertainty, which may pose challenges, as the signal of height increase is typically small. Other works usually simplify the problem by reducing change to a binary predicate (e.g., loss/no loss \cite{fogel_open-canopy_2024}), ignoring the continuous nature of height changes where both incremental growth and partial canopy loss are essential variables. Hence, robust and large-scale benchmark datasets capturing continuous height change remain scarce, constraining efforts to evaluate the accuracy of height models and remote sensing-based datasets. To our knowledge, no existing public benchmark dataset provides continuous canopy height change with change uncertainty over a large region and at high spatial resolution.\par
Here, we present the Canopy Height Change (CHC) dataset (fig. \ref{fig:overview}), a benchmark dataset for continuous canopy height change at \SI{3}{m} resolution on an area of \SI{10598}{\square\kilo\meter} in northern and western Spain. The changes, including continuous values on loss and gain, are derived from two national ALS campaigns from 2018 and 2023 \cite{cnig_lidar_2021, ign_lidar_2025} and associated with a simulated spatially resolved change uncertainty, using ALS acquisition geometry and sampling density to model systematic and statistical error propagation. We paired the change data with co-located time series of PlanetScope optical imagery between 2018 and 2023. This enables models to harness spatial and temporal patterns in the imagery to predict canopy height change.\par
We apply the CHC dataset to benchmark Geospatial Foundation Models (GFMs). Large-scale, pretrained GFMs have increasingly been adopted in deep learning and Earth observation \cite{huang_survey_2025}. In contrast to traditional remote sensing models, typically optimized for a single product and sensor, GFMs are designed to encode broad, semantically rich feature representations of land surface processes, which can then be fine-tuned with comparatively small datasets representing the target domain, such as canopy height \cite{tolan_very_2024}. The models are typically trained in a self-supervised approach on a wide range of multi-sensor Earth observation data, such as from optical and radar at various resolutions. Because it requires temporal reasoning, cross sensor transfer, and fine-scale spatial sensitivity, the CHC dataset represents a relevant use case for evaluating GFMs. The dataset is compatible with the PANGAEA benchmark framework \cite{marsocci_pangaea_2025} and can be accessed here: \url{https://sid.erda.dk/sharelink/eP4ENGhKTv}.\par
The main contributions of this study are as follows: 
\begin{enumerate}
    \renewcommand{\labelenumi}{(\theenumi)}
    \item A new CHC dataset, covering \SI{10598}{\square\kilo\meter} with \SI{3}{\meter} resolution labels of continuous height change with uncertainty quantification, and PlanetScope satellite imagery time series across 6 years.
    \item Accurate change uncertainty assessment during training and evaluation through quantified systematic and statistical data uncertainties based on simulation at pixel level.
    \item Performance analysis of state-of-the-art GFMs on the CHC dataset.
\end{enumerate}

\section{Related Work}
\label{sec:related}
\subsubsection*{Change detection datasets}
A large number of remote sensing change datasets exist for the benchmarking of change detection models in a variety of application domains. Typically, they combine manual change labels with co-located bi-temporal remote sensing imagery, between which the change is assessed. The change detection problem is usually classified into Binary Change Detection (BCD), Semantic Change Detection (SCD), and Remote Sensing Image Change Captioning (RSICC). Application domains span urban settings (construction and demolition)  (\eg, \cite{zhang_hyper-neighborhood_2025, zhou_segchange-r1_2025, wang_constructing_2025, liu_remote_2022}), land cover/land use change (\eg, \cite{liu_jl1-cd_2025, tuzlupinar_introducing_2025, tan_triples_2025}), and disaster assessments (\eg, \cite{wang_disasterm3_2025, hansch_spacenet_2022, zhang_cross-domain_2023}).

\subsubsection*{3D change detection.}
3D change detection adds an additional dimension, for example height, depth, or volume changes, where classic planimetric approaches might not be sufficient. The data is often sourced from ALS, Digital Elevation Models (DEMs), 3D models, or stereo/multi-view images and either represented as 3D difference (\eg, Euclidean), or as “2.5D”, when changes are projected onto a plane (\eg, difference between DEMs) \cite{qin_3d_2016}. Change detection from repeat ALS is used in a wide range of applications from coastlines to canopy structure \cite{okyay_airborne_2019}. However, public datasets that pair remote sensing imagery with ALS-based height change targets are rare (tab. \ref{tab:3dcd}).\par
In the context of urban areas, 3D changes are often DSM or point cloud-based \cite{de_gelis_change_2021}. The 3DCD dataset \cite{coletta_3dcd_2022, marsocci_inferring_2023} comprises DSM differences of artificial objects, such as from construction or demolition in a semantic change detection dataset of \SI{18.8}{\square\kilo\meter} area, combined with aerial imagery. Also in the urban domain, the SMARS dataset \cite{fuentes_reyes_2d3d_2023} provides synthetically generated DSM changes and optical imagery from simulated 3D scenes.  

\begin{table}
    \centering
    \caption{\textbf{ALS-based 3D change benchmark datasets}. CHC significantly increases spatial and temporal coverage compared to existing datasets, and introduces height uncertainty estimates.}
\begin{tabular}{l cccccccc}
\toprule
     & Domain & \makecell{Nb. \\ imgs.} & \makecell{Area \\ (km$^2$)} & Sensor 
     & \makecell{GSD \\ (m)} & Task & Uncertainty \\
     \midrule
     3DCD \cite{coletta_3dcd_2022} & Urban & 2 & 18.8 & Aerial & 0.5 & Continuous & $\times$ \\
    SMARS \cite{fuentes_reyes_2d3d_2023} & Urban & 2 & 12 & Synthetic & 0.3--0.5 & Ternary & $\times$ \\
    OpenCanopy-$\Delta$ \cite{fogel_open-canopy_2024} & Natural & 2 & 166 & SPOT 6--7 & 1.5 & Binary & $\times$ \\
    CHC (ours) & Natural & 6 & 10.6k & PlanetScope & 3.7 & Continuous & \checkmark \\
    \bottomrule
\end{tabular}
\label{tab:3dcd}
\end{table}

Open-Canopy \cite{fogel_open-canopy_2024} is an ALS-based tree height dataset at \SI{1.5}{\meter} spatial resolution, paired with SPOT 6-7 imagery. A subset of the data provides binary change between 2022 and 2023 on an area of \SI{166}{\square\kilo\meter}. However, only significant height reductions of more than \SI{15}{\meter} and of areas larger than \SI{200}{\square\meter} are included in the dataset. To our knowledge, no metric vegetation height dataset that continuously quantifies tree growth has been published so far.

\subsubsection*{Uncertainty of ALS-based canopy height.}
Discrete airborne LiDAR (Aerial Laser Scanning, ALS) is widely used to generate 3D representations of entire landscapes. Mounted on an aircraft, the LiDAR scanner can generate dense wall-to-wall point clouds with high accuracy, which in turn are used to create Digital Surface Models (DSMs) and Digital Terrain Models (DTMs) through rasterization \cite{coops_modelling_2021}. Uncertainties in LiDAR measurements can originate from several sources: positional errors include the horizontal and vertical uncertainty of points induced by uncertainty in the GPS and inertial measurement units and the angular and range accuracy of the sensor. Uncertainties in surface representation are created by the discrete sampling process of rough and complex surfaces, leading to occlusion and undersampling. Finally, the point cloud classification (\eg, into vegetation, buildings, infrastructure, etc.) introduces uncertainty depending on the classification algorithms applied \cite{glennie_rigorous_2007, passalacqua_analyzing_2015}.\par
In a multi-temporal setting, it is usually assumed that changes can be detected despite varying acquisition parameters, or that  the data is consistent across collections \cite{goulden_uncertainty_2017}. However, repeated ALS acquisitions typically come from different sensors and flight configurations. Variation in pulse density, driven by flight and sensor parameters, affects both DTM and DSM accuracy, and vertical and horizontal uncertainties must be evaluated across swaths \cite{sampath_geometric_2016}. 
Multivariate analyses have shown that scan geometry is a key driver of LiDAR height uncertainty and bias, with measured height, local variability of neighboring cells, and pulse density among the strongest predictors \cite{fradette_method_2019}. Additional influences include beam divergence, peak pulse power, canopy structure, terrain slope, and canopy edges \cite{goulden_uncertainty_2017, gopalakrishnan_prediction_2015}.
%
ALS systematically underestimates canopy height because the highest crown points are often missed \cite{fradette_method_2019, gatziolis_challenges_2010, hirata_effects_2004, roussel_removing_2017, wing_applying_2010, yu_evaluating_2024}, which is exacerbated by lower point densities \cite{zhao_utility_2018, zhao_domain-adaptive_2021}. Furthermore, systematic biases have been attributed to DSM algorithm choices, pulse penetration characteristics, and understory structure \cite{hyyppa_review_2008}. Species‑related traits such as crown shape also influence under/overestimation, and reduced pulse density (\eg, through point cloud thinning) exacerbates these species‑dependent biases \cite{sibona_direct_2017}.\par
Bias mitigation in LiDAR‑derived canopy heights typically relies on statistical correction models for varying acquisition conditions. Several approaches have been tested, although their explanatory power often remains modest. For example, \cite{fradette_method_2019} evaluated linear and non‑linear formulations to correct height underestimation and found limited improvements, reflecting the probabilistic nature of the discrete sampling. \cite{zhao_utility_2018} proposed to correct density-induced biases through modeling the bias-density relationship by thinning high-density samples to lower densities to capture a parameterized relation, noting diminishing biases above roughly \SI{7}{\square\ppm} (points per square meter). A probabilistic framework has also been introduced by \cite{roussel_removing_2017}, who modeled the stand‑level vertical forest structure and simulated the sampling process as a function of pulse density and footprint size.
\subsubsection*{Geospatial Foundation Models.}
Geospatial foundation models (GFMs) have become ubiquitous in recent years, challenging the established paradigm of fully-supervised learning for remote sensing tasks \cite{huang_survey_2025}. Powered by advances in self-supervised learning, they serve the purpose of generalistic feature extractors that can be applied to any modality and for a variety of downstream tasks. The models are typically trained on large, diverse remote sensing datasets through contrastive learning \cite{liu_remoteclip_2024}, generative learning \cite{reed_scale-mae_2023,xiong_neural_2025}, or supervised learning \cite{bastani_satlaspretrain_2023}. However, their limitations remain poorly understood due to their recency and bias towards classification tasks for which labels are more readily available \cite{yang_survey_2025}. On the PANGAEA benchmark \cite{marsocci_pangaea_2025}, which only includes two regression tasks (BioMassters \cite{nascetti_biomassters_2023} and Open-Canopy \cite{fogel_open-canopy_2024}), it has been noted that regression remains challenging, with supervised baselines still outperforming complex self-supervised GFMs.
\subsubsection*{Heteroscedastic regression.}
Heteroscedastic regression refers to supervised learning settings in which target noise varies across data points \cite{mai_batch_2021}. Although deep neural networks can tolerate a high degree of label noise when provided with a sufficiently large set of clean labels \cite{arpit_closer_2017, rolnick_deep_2018}, they also exhibit high capacity in fitting random noise, which can degrade performance under uncertain labels \cite{chen_understanding_2019, zhang_understanding_2017}. Various approaches have been explored to mitigate the impact of noisy labels \cite{song_learning_2022}. For regression, re-weighting the mean squared error (MSE) loss by target uncertainty can outperform a standard MSE by reducing the contribution of noisy training samples \cite{sai_learning_2009, mai_batch_2021}.

\vspace{5mm}
\section{Dataset}
\label{sec:dataset}
The CHC dataset is designed to assess a model’s capability to regress the metric change in canopy height, both decrease and increase, and hence its ability to detect tree removal and growth. In this section, we describe the underlying data, the processing, and the characteristics of the final benchmark dataset.

\subsection{General characteristics}
\begin{figure}[t]
    \centering
    \includegraphics[width=0.9\textwidth]{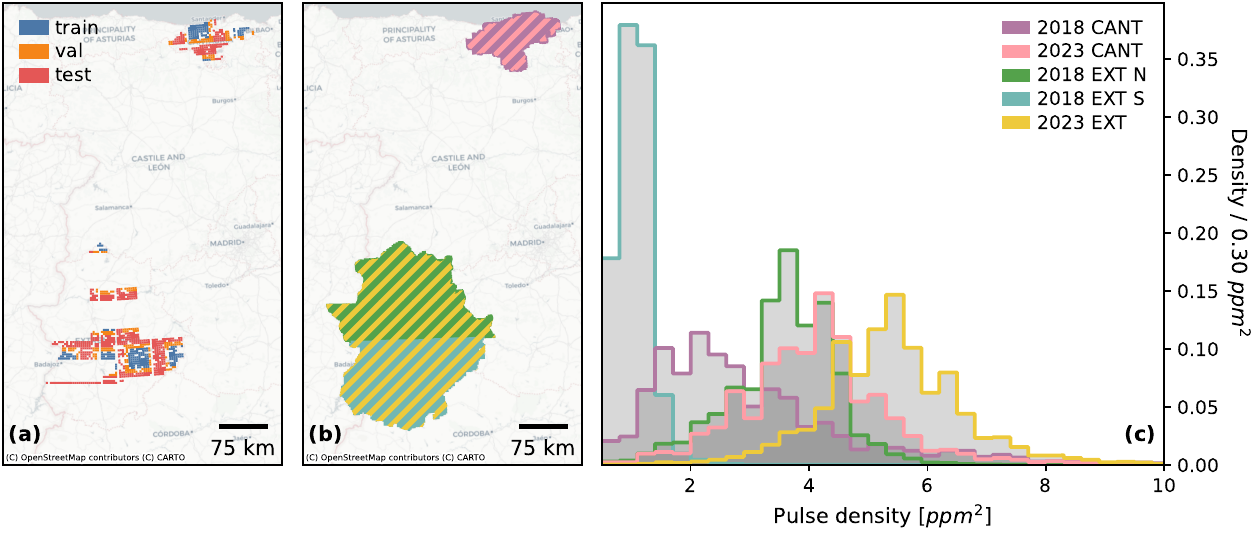}

    \caption{Overview of the dataset area of interest. (a) Geographic split of the dataset tiles in training, validation, and test set, organized in concentric squares, based on the method of \cite{sumbul_bigearthnet_2019}. The tiles only cover parts of the entire campaign area due to mismatches between flight seasons. (b) Selected PNOA ALS campaigns and overlaps between 2018 and 2023 for the regions of Cantabria (CANT) and Extremadura (EXT) North and South. (c) Distribution of pulse densities for each of the ALS campaigns in points per square meter (\si{\square\ppm}). The 2018 campaigns exhibit median pulse densities considerably lower than the 2023 campaigns, which can lead to uncertainty in the observed height changes. }
    \label{fig:map_densities}
\end{figure}

We made use of the publicly available point cloud data from the Spanish National Plan of Air Orthophotography (PNOA), conducted by the Insituto Geográfico National (IGN) \cite{cnig_lidar_2021, ign_lidar_2025}. The ALS data was collected in different nation-wide campaigns, out of which \textit{Cobertura 2a} (2015-2021) and \textit{3a} (2022-2026) fall into our observation period, as defined by the availability of the PlanetScope data. The measurements are organized in regional lots with varying flight dates, sensors, and pulse densities (fig. \ref{fig:map_densities}), depending on region and time.\par
We selected the areas that had double LiDAR coverage from both campaigns, leaving us with measurements from 2018 (\textit{2a}) and 2023 (\textit{3a}) between which we calculated change. Furthermore, we only selected areas covered during leaf-on period between May and October, to avoid cross-seasonal effects originating from the absence of leaves in one of the campaigns. The filtering resulted in an area of \SI{10598}{\square\kilo\meter}, covering the region Cantabria in the north of the country, and parts of Extremadura at the western border to Portugal. We used processed point clouds in format LAS v1.4, as delivered by CNIG \cite{cnig_lidar_2021, ign_lidar_2025} as input for the subsequent processing.\par
We paired the data with PlanetScope optical imagery from the same regions. The imagery is available with approximately \SI{3}{\meter} resolution at daily frequency, and RGB and Near-Infrared bands. We selected cloud-free scenes within a seasonal window based on the MODIS phenology product \cite{friedl_mcd12q2_2019} in late summertime, as done by \cite{liu_overlooked_2023}. If no coverage was reached at that time, we expanded the window until at least one image matched the criteria.\par
\subsection{Dataset creation}
\subsubsection*{Rasterization.}
We rasterized the LiDAR point clouds to digital surface models (DSM) by calculating the 95\textsuperscript{th} height percentile on a regular $3\times 3 \si{\meter}$ grid for each of the two years. The points were filtered based on the encoded ASPRS classes \cite{asprs_specification_2019}, and only the vegetation-related classes (\ie, classes 3-5) were included. Through vertical differencing of the DSMs, we computed the surface height change $\Delta h^{(95)}$. Compared to calculating the difference between canopy height models (CHMs), which represent the height above ground, this approach has the advantage of avoiding additional errors introduced by DTM measurements and interpolation. We masked areas where vegetation was below \SI{3}{m} in both years, to avoid impacts of non-woody vegetation changes, \eg field harvests. We calculated the explanatory features on the same grid, using 2D binned statistics (bin size was equal to GSD, \ie, \SI{3}{\meter}).
\subsubsection*{Uncertainty and offset estimation.}
DSM measurements contain both systematic and statistical uncertainties, which hinder the direct interpretation of observed height differences. Systematic uncertainty $\sigma_\mathrm{syst}$ typically originates from technical constraints and flight parameters, whereas statistical uncertainty $\sigma_\mathrm{stat}$ arises from the probabilistic process of measuring height at discrete locations through the LiDAR pulses and aggregating them into grid cells.\par
We estimated uncertainties and offsets between the campaigns through simulation and by analyzing the residuals between ALS campaigns at static locations. We estimated $\sigma_\mathrm{syst}$ from the residuals between the two ALS campaigns on flat road surfaces that we assumed remained unchanged throughout the observation period, based on the Spanish Land Cover and Land Use Information System (SIOSE) (IGN, 2015). We obtained a distribution of residuals with standard deviations between \SI{0.17}{\meter} and \SI{0.23}{\meter}, and offsets between \SI{-0.09}{\meter} and \SI{-0.01}{\meter}, depending on the combination of ALS campaigns.\par
In contrast, $\sigma_\mathrm{stat}$ is heteroscedastic and originates from the discrete LiDAR point measurements of complex geometric surfaces, and the aggregation of their 95\textsuperscript{th} height percentile ($h^{\left(95\right)}$) into grid cells. We assumed that grid cells $i$ with high pulse density ($\rho_i\geq \SI{10}{\square\ppm}$, which equals 90 pulses in the $3 \times 3$ grid cells) represent the surface geometry sufficiently well and used their heights ($h_i^{\left(95\right)}\left(\rho_i^{\left(10\right)}\right)$) as reference values. We then simulated the sampling process with $j$ lower pulse densities ($\SI{1}{\square\ppm}<\rho_i^{\left(j\right)}<\SI{10}{\square\ppm}$) and calculated the corresponding height difference $d_i^{\left(j\right)}=h_i^{\left(95\right)}\left(\rho_i^{\left(j\right)}\right)-h_i^{\left(95\right)}\left(\rho_i^{\left(10\right)}\right)$, along with other explanatory features $\mathbf{v}_i\left(\rho_i^{\left(j\right)}\right)$ that represent surface geometry and ALS acquisition geometry (tab. S2). We trained a multi-layer perceptron (MLP) adjustment model $f(\rho, \mathbf{v})$ on \num{7.935e5} training grid cells to predict height offset $\hat{b}_i$ and variance $\hat{\sigma}_i^2$ based on pulse density and explanatory features. Using the Gaussian negative log likelihood (NLL) loss (eq. \ref{eq:nll}), the model is incentivized to represent every prediction as a Gaussian distribution conditional on the input data, allowing the model to attenuate the effect of uncertain targets \cite{kendall_what_2017}.
\begin{align}
    \mathcal{L}_{\mathrm{NLL}}=\frac{1}{N}\sum_{i=1}^{N}\left(\frac{1}{2}\exp{\left(-\log{\hat{\sigma}_i^2}\right)}\left(b_i-\hat{b}_i\right)^2+\frac{1}{2}\log{\hat{\sigma}_i^2}\right)
    \label{eq:nll}
\end{align}
Finally, we combined systematic and statistical uncertainties through variances summation for all pixels in both campaigns. To illustrate the effect of the adjustment model, we built a building-specific model using static buildings to represent unchanged complex geometries. The model attained an Expected Calibration Error (ECE) of \SI{0.21}{\meter} for the predicted residual standard deviation, though it generally underestimated uncertainty relative to the observed residuals. Further details on uncertainty and offset estimation are provided in the supplementary materials.
\subsubsection{Dataset overview.}
\begin{figure}[t]
    \centering
    \includegraphics[width=0.9\linewidth]{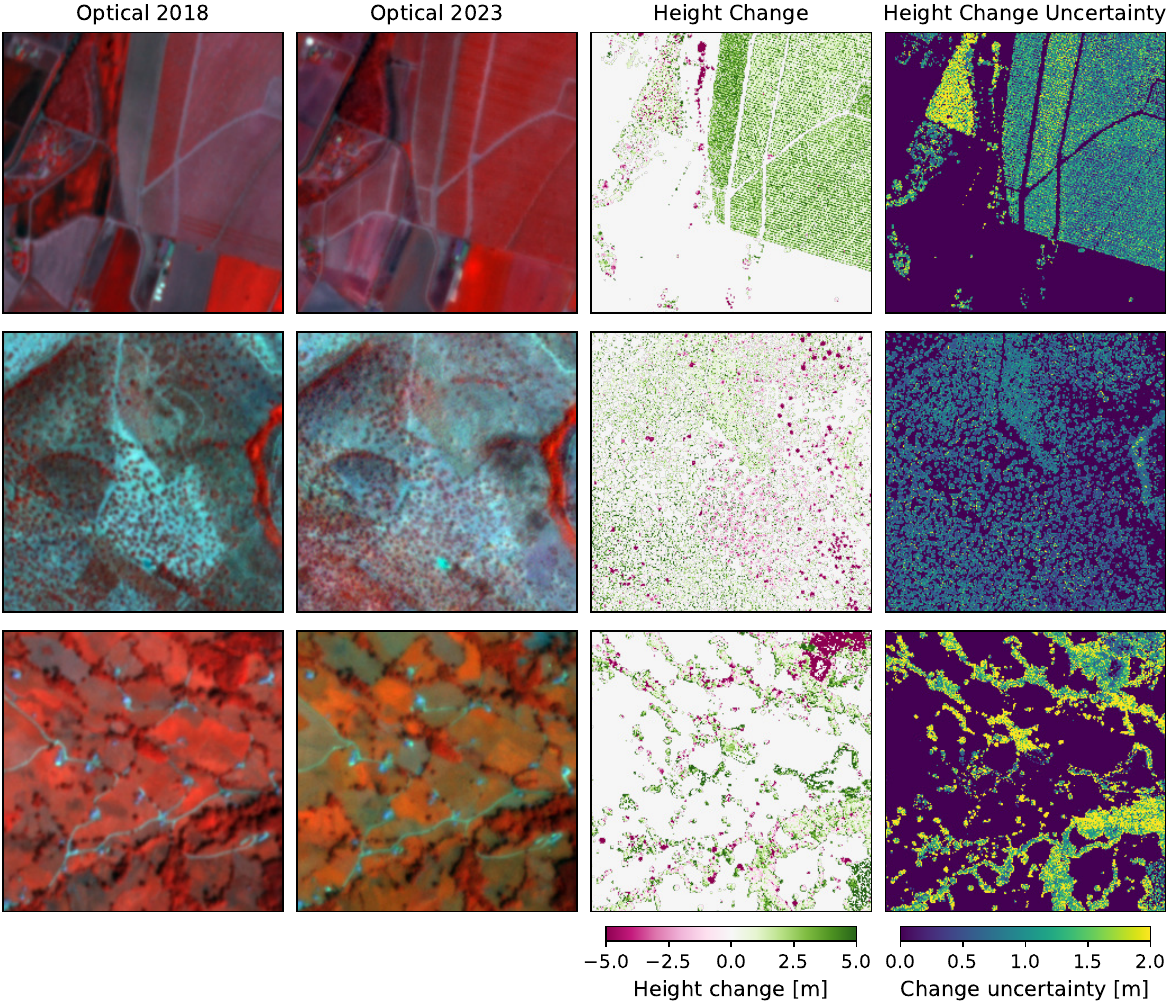}
    \caption{Examples of optical PlanetScope images from 2018 and 2023 (displayed as false color composites), the corresponding ALS-derived canopy height difference, and change uncertainty for an orchard, a silviopasture, and a grassland-forest mosaic (from top to bottom).}
    \label{fig:opt_diff_std}
\end{figure}
\begin{figure}[ht]
    \centering
    \includegraphics[width=0.9\linewidth]{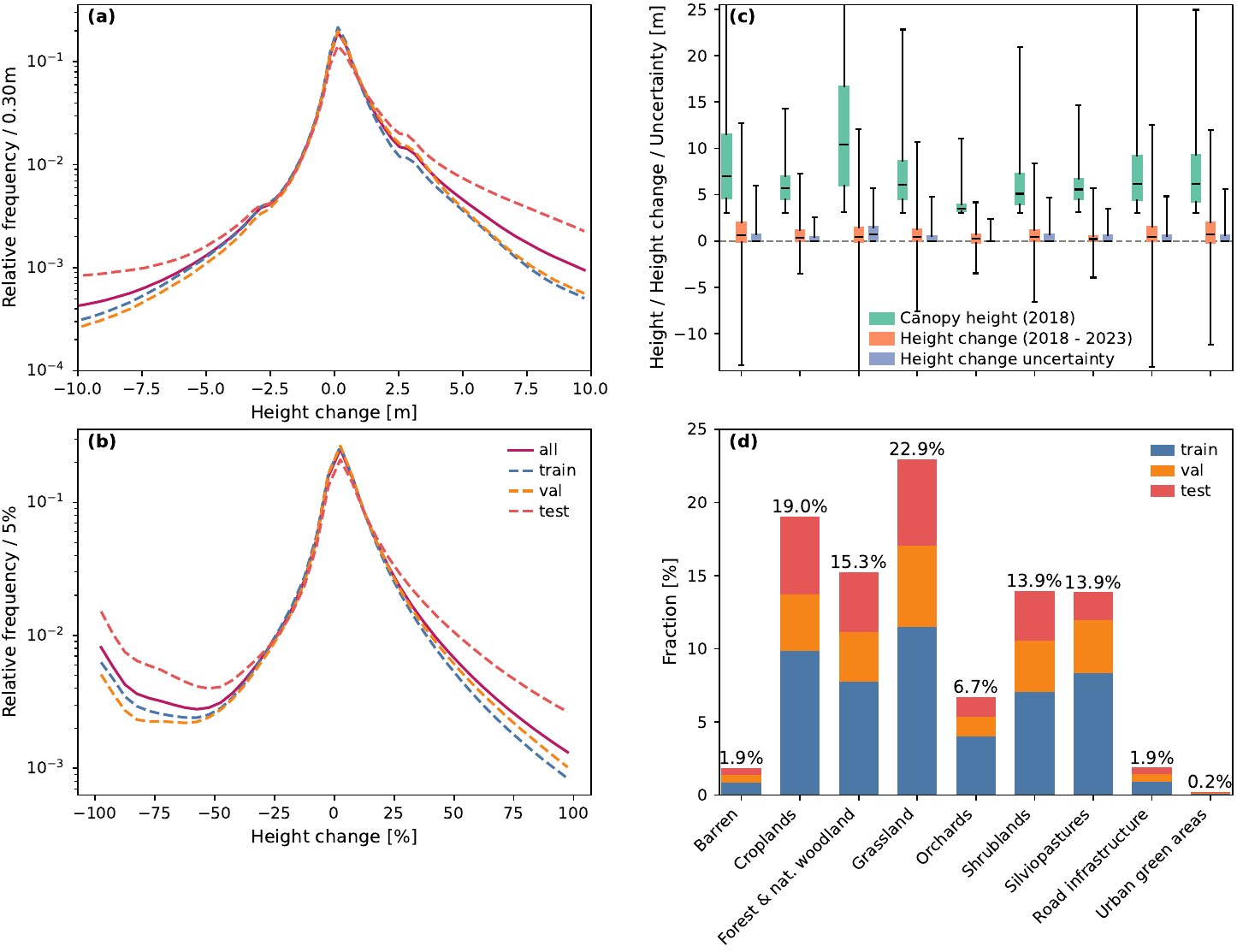}
    \caption{Overview of dataset statistics: (a) Absolute and (b) relative height change distribution for vegetation taller than \SI{3}{\meter} for each dataset split (bin width \SI{0.3}{\meter} and \SI{5}{\percent}, respectively); (c) Distribution of canopy height (2018, min \SI{3}{\meter} height), canopy height change (2018 – 2023), and canopy height change uncertainty for different SIOSE (2014) land cover classes. The whiskers represent the 1\textsuperscript{st} and 99\textsuperscript{th} percentile; (d) Distribution of the SIOSE (2014) land cover \cite{ign_spanish_2015} in the dataset area for each dataset split (unclassified landcover omitted).}
    \label{fig:stats}
\end{figure}

The dataset is organized in 1123 tiles of $1024\times 1024$ with \SI{3}{\meter} resolution, aligned with the PlanetScope imagery. PlanetScope is available as a time series between 2018 and 2023 with one to nine unordered images per year, allowing for averaging or temporal shift augmentation. The height-change tiles represent changes between 2018 and 2023 and have three data layers: height change (in \si{\meter}), maximum canopy height (height above ground) in both years (in \si{\meter}), and change variance (in \si{\square\meter}) (fig. \ref{fig:opt_diff_std}). We apply a threshold of min. \SI{3}{\meter} height to separate trees from other vegetation. 
\SI{27}{\percent} of the area is covered with vegetation higher than \SI{3}{\meter} in at least one of the years. The dataset covers a variety of land cover types, with Grassland, Croplands, and Forests being the most prominent ones (fig. \ref{fig:stats}d).
The canopy height shows a long tail distribution and is on average slightly lower in the train set compared to validation and test set. The height changes at vegetated locations are centered around zero, steeply sloping on both sides of the distribution (fig. \ref{fig:stats}a). The relative changes show a similar pattern, while a local peak at \SI{-100}{\percent} represents complete tree removal. These changes represent raw changes and might be obscured by non-significant variations in canopy height, which can occur, \eg, at canopy edges.\par

We observe different median canopy heights in the various land cover classes represented in the dataset. Forest and natural woodlands show the highest canopy (median \SI{10.4}{\meter}) and orchards the lowest (median \SI{3.5}{\meter}), just above the minimum height threshold (fig. \ref{fig:stats}c). Note that non-forest land cover can still include scattered trees that do not meet the criteria for forests. The median changes are positive in all land cover classes, ranging from median increases of \SI{0.2}{\meter} in silviopastures to \SI{0.7}{\meter} in urban green areas, which, however, might be confounded by misclassified buildings in the LiDAR point clouds. The largest median height change uncertainty is found for forests and natural woodlands (\SI{1.1}{\meter}), the lowest for orchards (\SI{0.8}{\meter}).

\subsubsection{Splits.}

The dataset is split (fig. \ref{fig:map_densities}) into training (592 tiles), validation (262 tiles) and test set (269 tiles) by region in concentric squares with the validation set acting as a buffer between train and test set \cite{sumbul_bigearthnet_2019}. The test split only contains low-uncertainty change values, we applied a vegetation mask and retained only pixels with an absolute change $z$-score above 1.65, corresponding to cases where the observed deviation would occur with less than \SI{10}{\percent} probability under a no-change scenario.
\section{Experiments}
\label{sec:experiments}
We used the CHC dataset to assess the performance of GFMs on the task of canopy height change regression. We employed pretrained GFM encoders and finetuned only the decoder on the training split. Model performance was assessed on the heldout test split, examining several approaches for integrating label uncertainty during training. 
\subsection{Experimental setting}
\subsubsection*{Competing methods.} \textbf{Scale-MAE} \cite{reed_scale-mae_2023} is a scale‑aware autoencoder that learns multiscale geospatial features using a ViT encoder with Ground Sample Distance positional encoding and a Laplacian‑pyramid decoder. It is pretrained via masked image modeling on multiscale remote‑sensing data. \textbf{RemoteCLIP} \cite{liu_remoteclip_2024} adapts CLIP-style dual encoders for remote sensing by jointly training image and text encoders using largescale image-caption data, including UAV imagery. This enables learning semantically aligned visual-text representations suitable for zeroshot classification and other downstream tasks. \textbf{DOFA} \cite{xiong_neural_2025} is designed to handle inputs from diverse remote sensing sensors. It uses a wavelength conditioned dynamic hypernetwork to generate modality adaptive parameters, allowing for arbitrary input channel counts. DOFA is pretrained on five heterogeneous modalities for cross sensor generalization and robust performance on unseen modalities. \textbf{Prithvi-EO 2.0} \cite{szwarcman_prithvi-eo-20_2025} is a temporal ViT foundation model pretrained using a masked autoencoder objective on Harmonized Landsat–Sentinel2 (HLS) data. It uses inter-patch spatial attention and intra-patch temporal attention, enabling spatiotemporal modeling of satellite time series. \textbf{CROMA} \cite{fuller_croma_2023} uses contrastive learning with masked autoencoding to exploit spatially aligned optical and radar imagery from Sentinel-1 and 2. It integrates separate unimodal encoders for optical and radar inputs (contrastive learning) and a joint multimodal encoder (reconstruction loss). \textbf{SatlasNet} \cite{bastani_satlaspretrain_2023} is trained on a large collection of remote‑sensing datasets spanning segmentation, regression, detection, and classification tasks. It processes time‑series inputs with Swin‑Transformer backbones and fuses features via temporal max‑pooling at multiple scales. \textbf{SpectralGPT} \cite{hong_spectralgpt_2024} uses masked autoencoding to learn visual representations in a progressive manner, using various Sentinel-2-based data sources and a ViT-based backbone architecture. The fine-tuning includes downstream tasks such as image classification, segmentation, and change detection. \textbf{TerraMind} \cite{jakubik_terramind_2025} uses modality specific ViT tokenizers to embed data from various sources, such as Sentinel-1 and 2, land cover maps, or image captions. During pretraining, the model learns to reconstruct missing modalities through a masked modeling approach. \textbf{DINOv3} \cite{simeoni_dinov3_2025} is a vision transformer pretrained through self-distillation, producing dense features that have been shown to generalize well across remote sensing tasks. \textbf{ECHOSAT} \cite{pauls_capturing_2025} is a global CHM time series dataset at \SI{10}{\meter} resolution, generated with a spatiotemporal transformer model that performs pixel-wise regression of canopy height from optical and radar imagery, using a growth-constrained loss and GEDI-based canopy height to enforce physically realistic forest dynamics over time.
\subsubsection*{Training and evaluation setup.}
We evaluated the performance of the GFMs using the PANGAEA benchmark framework \cite{marsocci_pangaea_2025}. Each model received time series of annual PlanetScope imagery as input and predicted canopy height change in a dense prediction task. We used the frozen pre-trained GFM encoders to create embeddings at different depths of the encoder, which we then assembled into feature pyramids. These were passed on to a trainable UPerNet decoder \cite{xiao_unified_2018} which performed multi-scale aggregation and produced dense predictions of canopy height change. We upscaled the output features with bilinear interpolation and added a residual refinement block at the end. 
\par
The encoders were natively pre-trained following different objectives and data regimes. Hence, not all encoders were compatible in terms of required input channels or time-series capacities. PANGAEA aligns input channels with those used to train the GFM, zero‑padding any unmatched channels \cite{marsocci_pangaea_2025}. For encoders without native support of time series input, PANGAEA introduces an intermediate Lightweight Temporal Attention Encoder (LTAE) \cite{garnot_lightweight_2020} after the encoder. Each time step is embedded individually and then passed to the LTAE, which applies attention with temporal position cues to fuse these features across time steps. The fused feature pyramids are then passed to the UPerNet decoder.
\begin{align}
    \mathrm{wMSE}=\frac{\sum_{i=1}^{N}\left[\frac{1}{\sigma_i^2}\left(\hat{h}_i-h_i\right)^2\right]}{\sum_{i=1}^{N}\frac{1}{\sigma_i^2}}
    \label{eq:wmse}
\end{align}
The training of the learnable UPerNet decoder parameters was performed for each encoder separately over 50 epochs on the training split of the CHC dataset, with AdamW optimizer, learning rate warmup, and cosine scheduler. We applied a vegetation mask to the dataset to only train and validate at locations with vegetation higher than \SI{3}{\meter}. The mask was eroded by one pixel to remove confounding effects on vegetation edges.\par
We compared two loss‑weighting schemes to assess how GFMs behave with and without pixel‑level uncertainty. The weighted MSE (wMSE) \cite{mai_batch_2021} (eq. \ref{eq:wmse}) scales errors by the inverse target variance, down‑weighting uncertain height‑change estimates so the model focuses on more reliable signals. The thresholded MSE (tMSE) instead computes the loss only for pixels with a z‑score above 1.65, excluding low‑signal‑to‑noise samples. Finally, the standard MSE treats all targets as equally reliable. We then evaluated the models on the test split of the CHC dataset, using root mean squared error (RMSE), mean absolute error (MAE), normalized MAE (nMAE), and coefficient of determination ($R^2$). In addition to overall metrics, we separately masked increases and decreases to assess performance on positive and negative height changes.\par
In addition to the GFMs, we included a UNet \cite{ronneberger_u-net_2015} and a ResNet-50 \cite{he_deep_2015} with UPerNet decoder as baseline models that were trained fully supervised 
without any frozen parameters. We also included a randomly initialized ResNet-50 encoder that was not updated during training, as a null model. Training and evaluation followed the same logic as for the GFMs. Furthermore, the ECHOSAT dataset \cite{pauls_echosat_2026} acted as an independent baseline trained globally baselineon GEDI-derived canopy height. 
\subsection{Results}
\subsubsection{Performance of the benchmarked GFMs}
The performance of the evaluated baselines and GFMs on the CHC test split showed that none of the GFMs could outperform a supervised UNet trained with the thresholded MSE (tMSE) loss, which achieved an RMSE of \SI{5.65}{\meter} (tab. \ref{tab:model_metrics}). Prithvi and DOFA achieved the best overall results among the GFMs, with an RMSE of \SI{5.90}{\meter} and \SI{5.99}{\meter}, respectively. On the other hand, Scale-MAE was inferior to the baseline of a randomly initialized and non-fine-tuned ResNet-50 encoder across all loss weighting schemes. The ECHOSAT dataset outperformed both GFMs and supervised baselines with an RMSE of \SI{5.25}{\meter}. The results suggest that the fine-grained, subpixel structural cues driving continuous height change remain challenging for all of the frozen encoders when used for regression on PlanetScope. Moreover, task-specific models, such as the ECHOSAT Temporal-Swin-UNet trained on GEDI-derived canopy height, show superior performance, despite the coarser spatial resolution of \SI{10}{\meter}. Current remote sensing GFMs often struggle to outperform supervised baselines and specialized models on pixel level tasks \cite{marsocci_pangaea_2025, xu_specialized_2025, mai_on_2024}.\par

\begin{table}
    \centering
    \caption{Benchmark results of the GFMs on the CHC dataset, broken down in height increase, height decrease and all directions categories. We compare models trained with a standard MSE loss (s), uncertainty-thresholded version (t) and uncertainty-weighted (w), each in their best performing setup (see tab. S5 for the complete evaluation). Best performing models are highlighted in bold, and metrics equal or worse than a frozen and randomly initialized ResNet-50 encoder are greyed out.}

    \resizebox{\linewidth}{!}{
\begin{tabular}{l ccc | cccc cccc cccc}
 \toprule
 \multirow{2}{*}{Encoder} &
\multicolumn{3}{c}{\multirow{2}{*}{MSE}}
 & \multicolumn{4}{c}{\small Increase}
 &
 \multicolumn{4}{c}{\small Decrease}
 &
 \multicolumn{4}{c}{\small All}\\
 \cmidrule(lr){5-8}
 \cmidrule(lr){9-12}
 \cmidrule(lr){13-16}
 & s & t & w &
\small RMSE $\downarrow$ & \small MAE $\downarrow$ & \small nMAE $\downarrow$ & \small \(R^2\) $\uparrow$ 
&
\small RMSE $\downarrow$ & \small MAE $\downarrow$ & \small nMAE $\downarrow$ & \small \(R^2\) $\uparrow$ 
&
\small RMSE $\downarrow$ & \small MAE $\downarrow$ & \small nMAE $\downarrow$ & \small \(R^2\) $\uparrow$ \\
\midrule
        ResNet-50 & & & x & 3.51 & 2.57 & 0.68 & -1.05 & 11.47 & 9.05 & 0.94 & -1.09 & 6.98 & 4.55 & 0.82 & 0.21 \\
        UNet & & & x & 4.11 & 2.95 & 0.78 & -1.81 & 8.13 & 6.03 & 0.63 & -0.05 & 5.65 & 3.89 & 0.70 & 0.48 \\
        ECHOSAT & & & & 3.81 & 2.83 & 0.75 & -1.43 & \textbf{7.55} & \textbf{5.68} & \textbf{0.59} & \textbf{0.09} & \textbf{5.25} & \textbf{3.70} & \textbf{0.67} & \textbf{0.55}\\
        \midrule
        CROMA & & & x & \textbf{3.22} & \textbf{2.33} & \textbf{0.62} & \textbf{-0.73} & 10.94 & 8.91 & 0.93 & -0.90 & 6.61 & 4.34 & 0.78 & 0.29 \\
        DINOv3 & & x & & 4.16 & 3.19 & 0.85 & -1.89 & 8.76 & 6.49 & 0.68 & -0.22 & 5.95 & 4.20 & 0.76 & 0.42\\
        DOFA & & & x & 4.53 & 3.17 & 0.84 & -2.42 & 8.42 & 6.35 & 0.66 & -0.13 & 5.99 & 4.14 & 0.75 & 0.41 \\
        Prithvi & & x & & 4.27 & 3.44 & 0.91 & -2.05 & 8.52 & 6.34 & 0.66 & -0.15 & 5.90 & 4.32 & 0.78 & 0.43 \\
        RemoteCLIP & & & x & \textcolor{gray}{5.39} & \textcolor{gray}{4.02} & \textcolor{gray}{1.06} & \textcolor{gray}{-3.84} & 10.00 & 7.57 & 0.79 & -0.59 & 7.12 & 5.10 & 0.92 & 0.17 \\
        SatlasNet &  & x & & \textcolor{gray}{4.97} & 3.66 & 0.97 & \textcolor{gray}{-3.12} & 8.90 & 6.83 & 0.71 & -0.26 & 6.43 & 4.63 & 0.83 & 0.33\\
        Scale-MAE & x & & & \textcolor{gray}{4.58} & \textcolor{gray}{3.87} & \textcolor{gray}{1.02} & \textcolor{gray}{-2.49} & 12.38 & \textcolor{gray}{9.51} & \textcolor{gray}{0.99} & \textcolor{gray}{-1.44} & \textcolor{gray}{7.83} & \textcolor{gray}{5.59} & \textcolor{gray}{1.01} & \textcolor{gray}{0.00} \\
        SpectralGPT+ & x & & & \textcolor{gray}{4.59} & 3.80 & 1.01 & \textcolor{gray}{-2.51} & 11.11 & 8.61 & 0.90 & -0.96 & 7.23 & 5.27 & 0.95 & 0.15 \\
        TerraMind & x & & & 4.40 & 3.63 & 0.96 & -2.23 & 11.15 & 8.59 & 0.90 & -0.97 & 7.17 & 5.14 & 0.93 & 0.16 \\
 \bottomrule
  \end{tabular}}
    \label{tab:model_metrics}
\end{table}

\begin{table}[tb]
    \centering
    \caption{Benchmark results for predicting the direction of change. The F1-scores of the best performing models are highlighted in bold and scores equal or worse than the frozen and randomly initialized ResNet-50 encoder are greyed out (see tab. S6 for the complete evaluation). ResNet-50 and UNet models are trained fully supervised and serve as a baseline. All other models have frozen encoders, and only the decoder weights are trained.}
    
\begin{tabular}{l ccc | ccc ccc}
 \toprule
 \multirow{2}{*}{Encoder} &
\multicolumn{3}{c}{\multirow{2}{*}{MSE}}
 & \multicolumn{3}{c}{\small Increase}
 &
 \multicolumn{3}{c}{\small Decrease}\\
 \cmidrule(lr){5-7}
 \cmidrule(lr){8-10} & s & t & w &
\small Prec. & \small Recall & \small F1
&
\small Prec. & \small Recall & \small F1\\
\midrule
        ResNet-50 & x & & & 0.79 & 0.71 & 0.75 & 0.46 & 0.57 & 0.51 \\
        ResNet-50 & & x & & 0.76 & 0.94 & 0.84 & 0.70 & 0.34 & \textcolor{gray}{0.45} \\
        UNet & & x & & 0.85 & 0.80 & 0.83 & 0.60 & 0.69 & \textbf{0.64} \\
        ECHOSAT & & & & 0.83 & 0.85 & 0.84 & 0.63 & 0.59 & 0.61\\
        \midrule
        CROMA & & x & & 0.77 & 0.96 & \textbf{0.85} & 0.79 & 0.33 & \textcolor{gray}{0.47} \\
        DINOv3 & & & x & 0.83 & 0.76 & 0.80 & 0.55 & 0.65 & 0.59\\
        DOFA & & x & & 0.83 & 0.77 & 0.79 & 0.54 & 0.63 & 0.58 \\
        Prithvi & & x & & 0.81 & 0.83 & 0.82 & 0.60 & 0.57 & 0.58 \\
        Prithvi & & & x & 0.84 & 0.79 & 0.81 & 0.57 & 0.65 & 0.61 \\
        RemoteCLIP & x & & & 0.69 & 0.41 & 0.51 & 0.30 & 0.59 & \textcolor{gray}{0.40} \\
        RemoteCLIP & & x & & 0.77 & 0.63 & 0.69 & 0.40 & 0.57 & \textcolor{gray}{0.47} \\
        SatlasNet & x & & & 0.69 & 0.99 & 0.82 & 0.28 & 0.01 & \textcolor{gray}{0.02} \\
        SatlasNet & & x & & 0.82 & 0.74 & 0.78 & 0.51 & 0.63 & 0.57 \\
        Scale-MAE & x & & & 0.69 & 0.00 & \textcolor{gray}{0.00} & 0.31 & 1.00 & \textcolor{gray}{0.47} \\
        Scale-MAE & & & x & 0.62 & 0.07 & 0.12 & 0.30 & 0.91 & \textcolor{gray}{0.45} \\
        SpectralGPT+ & x & & & 0.79 & 0.76 & 0.77 & 0.49 & 0.53 & 0.51 \\
        TerraMind & x & & & 0.78 & 0.88 & 0.82 & 0.61 & 0.42 & 0.50 \\
        TerraMind & & & x & 0.76 & 0.94 & 0.84 & 0.69 & 0.33 & \textcolor{gray}{0.44} \\
 \bottomrule
  \end{tabular}
    \label{tab:classification_metrics}
\end{table}

Furthermore, we investigated the models’ performance in spatially detecting the direction of change on the test set (tab. \ref{tab:classification_metrics}). For detecting height increases, CROMA outperformed the supervised baselines with an F1 score of 0.85, while a supervised UNet achieved highest performance for decreasing heights ($\mathrm{F1} = 0.64$). Several GFMs could not outperform the randomly initialized and non-fine-tuned ResNet-50 encoder in detecting decrease, such as CROMA, RemoteCLIP, and Scale-MAE.
\subsubsection*{Height change uncertainty}
We investigated if providing target uncertainty to the models at training time by using wMSE or tMSE improves their performance on the test split. The results show that the best-performing models benefit from uncertainty-based weighting or thresholding. When trained with the tMSE loss, the RMSE of DOFA and the baseline UNet decreased by \SI{4}{\percent} and \SI{9}{\percent}, respectively, compared to the same models trained with the MSE loss. Training Prithvi using wMSE decreased the RMSE by \SI{8}{\percent} on the test split. GFMs with lower overall performance show a less clear response to uncertainty weighting.
\section{Conclusion}
We presented the CHC canopy height change dataset, to our knowledge the first public benchmark dataset for continuous canopy height change regression with associated uncertainty. 
We benchmarked popular Geospatial Foundation Models (GFMs) on the task of height change regression and reported an overall low performance compared to a supervised UNet baseline and the task-specific ECHOSAT dataset. All models had difficulties regressing the change magnitude, while to some degree being able to estimate the direction of change. This indicates a challenging task, that we hope will foster research to address the notable gap in performance compared to classification tasks.

Notably, we showed that uncertainty estimates may play an important role, by (1) eliminating low-quality test samples in a fundamentally noisy task, and (2) providing an adjustment variable to tune the quality/quantity balance in the fine-tuning sets for GFMs. Our work provides essential data for measuring real canopy height change, and supports future research in environmental monitoring with remote sensing.

\section*{Acknowledgements}
RF and MB acknowledge funding from the Danish National Research Foundation, Center for Remote Sensing and Deep Learning of Global Tree Resources (TreeSense), DNRF192. We thank Planet Labs PBC for the provision of the PlanetScope imagery.

%
%
\bibliographystyle{splncs04}
\bibliography{ref}

\begin{thebibliography}{10}
\providecommand{\url}[1]{\texttt{#1}}
\providecommand{\urlprefix}{URL }
\providecommand{\doi}[1]{https://doi.org/#1}

\bibitem{arpit_closer_2017}
Arpit, D., Jastrz\k{e}bski, S., Ballas, N., Krueger, D., Bengio, E., Kanwal, M.S., Maharaj, T., Fischer, A., Courville, A., Bengio, Y., Lacoste-Julien, S.: A closer look at memorization in deep networks. In: ICML (2017)

\bibitem{asprs_specification_2019}
{ASPRS}: {LAS} specification 1.4 -- {R15} (2019), \url{https://paulbourke.net/dataformats/laz/LAS_1_4_r15.pdf}

\bibitem{bastani_satlaspretrain_2023}
Bastani, F., Wolters, P., Gupta, R., Ferdinando, J., Kembhavi, A.: {SatlasPretrain}: a large-scale dataset for remote sensing image understanding. In: ICCV (2023)

\bibitem{chen_understanding_2019}
Chen, P., Liao, B.B., Chen, G., Zhang, S.: Understanding and utilizing deep neural networks trained with noisy labels. In: ICML (2019)

\bibitem{cnig_lidar_2021}
{CNIG, Centro Nacional de Información}: {LIDAR} 2ª cobertura (2015--2021) (2021), \url{https://centrodedescargas.cnig.es/CentroDescargas/lidar-segunda-cobertura}

\bibitem{coletta_3dcd_2022}
Coletta, V., Marsocci, V., Ravanelli, R.: {3DCD}: a new dataset for {2D} and {3D} change detection using deep learning techniques. ISPRS Archives  (2022)

\bibitem{coops_modelling_2021}
Coops, N.C., Tompalski, P., Goodbody, T.R.H., Queinnec, M., Luther, J.E., Bolton, D.K., White, J.C., Wulder, M.A., van Lier, O.R., Hermosilla, T.: Modelling lidar-derived estimates of forest attributes over space and time: a review of approaches and future trends. Remote Sensing of Environment  (2021)

\bibitem{fogel_open-canopy_2024}
Fogel, F., Perron, Y., Besic, N., Saint-André, L., Pellissier-Tanon, A., Schwartz, M., Boudras, T., Fayad, I., d'Aspremont, A., Landrieu, L., Ciais, P.: Open-{Canopy}: towards very high resolution forest monitoring. In: CVPR (2025)

\bibitem{fradette_method_2019}
Fradette, M.S., Leboeuf, A., Riopel, M., Bégin, J.: Method to reduce the bias on digital terrain model and canopy height model from {LiDAR} data. Remote Sensing  (2019)

\bibitem{friedl_mcd12q2_2019}
Friedl, M., Gray, J., Sulla-Menashe, D.: {MCD12Q2} {MODIS/Terra+Aqua} land cover dynamics yearly {L3} global 500m {SIN} grid {V006} (2019), \url{https://www.earthdata.nasa.gov/data/catalog/lpcloud-mcd12q2-006}

\bibitem{fuentes_reyes_2d3d_2023}
Fuentes~Reyes, M., Xie, Y., Yuan, X., d'Angelo, P., Kurz, F., Cerra, D., Tian, J.: A {2D}/{3D} multimodal data simulation approach with applications on urban semantic segmentation, building extraction and change detection. ISPRS Journal of Photogrammetry and Remote Sensing  (2023)

\bibitem{fuller_croma_2023}
Fuller, A., Millard, K., Green, J.R.: {CROMA}: remote sensing representations with contrastive radar-optical masked autoencoders. In: NeurIPS (2023)

\bibitem{garnot_lightweight_2020}
Garnot, V.S.F., Landrieu, L.: Lightweight temporal self-attention for classifying satellite image time series. In: ECML-PKDD Workshop on Advanced Analytics and Learning on Temporal Data (AALTD) (2020)

\bibitem{gatziolis_challenges_2010}
Gatziolis, D., Fried, J.S., Monleon, V.S.: Challenges to estimating tree height via {LiDAR} in closed-canopy forests: a parable from western {Oregon}. Forest Science  (2010)

\bibitem{glennie_rigorous_2007}
Glennie, C.: Rigorous {3D} error analysis of kinematic scanning {LIDAR} systems. Journal of Applied Geodesy  (2007)

\bibitem{gopalakrishnan_prediction_2015}
Gopalakrishnan, R., Thomas, V.A., Coulston, J.W., Wynne, R.H.: Prediction of canopy heights over a large region using heterogeneous lidar datasets: efficacy and challenges. Remote Sensing  (2015)

\bibitem{goulden_uncertainty_2017}
Goulden, T., Hass, B., Leisso, N.: Uncertainty in lidar derived canopy height models in three unique forest ecosystems. In: IGARSS (2017)

\bibitem{de_gelis_change_2021}
de~Gélis, I., Lefèvre, S., Corpetti, T.: Change detection in urban point clouds: an experimental comparison with simulated {3D} datasets. Remote Sensing  (2021)

\bibitem{hansch_spacenet_2022}
Hansch, R., Arndt, J., Lunga, D., Gibb, M., Pedelose, T., Boedihardjo, A., Petrie, D., Bacastow, T.M.: {SpaceNet} 8 - the detection of flooded roads and buildings. In: CVPRW (2022)

\bibitem{he_deep_2015}
He, K., Zhang, X., Ren, S., Sun, J.: Deep residual learning for image recognition. In: CVPR (2016)

\bibitem{hirata_effects_2004}
Hirata, Y.: The effects of footprint size and sampling density in airborne laser scanning to extract individual trees in mountainous terrain  (2004)

\bibitem{hong_spectralgpt_2024}
Hong, D., Zhang, B., Li, X., Li, Y., Li, C., Yao, J., Yokoya, N., Li, H., Ghamisi, P., Jia, X., Plaza, A., Gamba, P., Benediktsson, J.A., Chanussot, J.: {SpectralGPT}: spectral remote sensing foundation model. IEEE Transactions on Pattern Analysis and Machine Intelligence  (2024)

\bibitem{huang_survey_2025}
Huang, Z., Yan, H., Zhan, Q., Yang, S., Zhang, M., Zhang, C., Lei, Y., Liu, Z., Liu, Q., Wang, Y.: A survey on remote sensing foundation models: from vision to multimodality. arXiv:2503.22081  (2025)

\bibitem{hyyppa_review_2008}
Hyyppä, J., Hyyppä, H., Leckie, D., Gougeon, F., Yu, X., Maltamo, M.: Review of methods of small-footprint airborne laser scanning for extracting forest inventory data in boreal forests. International Journal of Remote Sensing  (2008)

\bibitem{ign_spanish_2015}
{IGN}: Spanish land cover and land use system ({SIOSE}) 1:25.000 (2015), \url{https://www.siose.es/en/web/guest/descripcion}

\bibitem{ign_lidar_2025}
{IGN, Centro Nacional de Información}: {LIDAR} 3ª cobertura (2022--2025) (2025), \url{https://centrodedescargas.cnig.es/CentroDescargas/lidar-tercera-cobertura}

\bibitem{jakubik_terramind_2025}
Jakubik, J., Yang, F., Blumenstiel, B., Scheurer, E., Sedona, R., Maurogiovanni, S., Bosmans, J., Dionelis, N., Marsocci, V., Kopp, N., Ramachandran, R., Fraccaro, P., Brunschwiler, T., Cavallaro, G., Bernabe-Moreno, J., Long\'ep\'e, N.: Terramind: Large-scale generative multimodality for earth observation. In: ICCV (2025)

\bibitem{kendall_what_2017}
Kendall, A., Gal, Y.: What uncertainties do we need in {Bayesian} deep learning for computer vision? In: NeurIPS (2017)

\bibitem{lang_high-resolution_2023}
Lang, N., Jetz, W., Schindler, K., Wegner, J.D.: A high-resolution canopy height model of the {Earth}. Nature Ecology \& Evolution  (2023)

\bibitem{liu_remote_2022}
Liu, C., Zhao, R., Chen, H., Zou, Z., Shi, Z.: Remote sensing image change captioning with dual-branch transformers: a new method and a large scale dataset. IEEE Transactions on Geoscience and Remote Sensing  (2022)

\bibitem{liu_remoteclip_2024}
Liu, F., Chen, D., Guan, Z., Zhou, X., Zhu, J., Ye, Q., Fu, L., Zhou, J.: {RemoteCLIP}: a vision language foundation model for remote sensing. IEEE Transactions on Geoscience and Remote Sensing  (2024)

\bibitem{liu_overlooked_2023}
Liu, S., Brandt, M., Nord-Larsen, T., Chave, J., Reiner, F., Lang, N., Tong, X., Ciais, P., Igel, C., Pascual, A., Guerra-Hernandez, J., Li, S., Mugabowindekwe, M., Saatchi, S., Yue, Y., Chen, Z., Fensholt, R.: The overlooked contribution of trees outside forests to tree cover and woody biomass across {Europe}. Science Advances  (2023)

\bibitem{liu_jl1-cd_2025}
Liu, Z., Zhu, R., Gao, L., Zhou, Y., Ma, J., Gu, Y.: {JL1-CD}: a new benchmark for remote sensing change detection and a robust multi-teacher knowledge distillation framework. arXiv:2502.13407  (2025)

\bibitem{mai_on_2024}
Mai, G., Huang, W., Sun, J., Song, S., Mishra, D., Liu, N., Gao, S., Liu, T., Cong, G., Hu, Y., Cundy, C., Li, Z., Zhu, R., Lao, N.: On the opportunities and challenges of foundation models for {GeoAI} (vision paper). ACM Transactions on Spatial Algorithms and Systems  (2024)

\bibitem{mai_batch_2021}
Mai, V., Khamies, W., Paull, L.: Batch inverse-variance weighting: deep heteroscedastic regression. arXiv:2107.04497  (2021)

\bibitem{marsocci_inferring_2023}
Marsocci, V., Coletta, V., Ravanelli, R., Scardapane, S., Crespi, M.: Inferring {3D} change detection from bitemporal optical images. ISPRS Journal of Photogrammetry and Remote Sensing  (2023)

\bibitem{marsocci_pangaea_2025}
Marsocci, V., Jia, Y., Bellier, G.L., Kerekes, D., Zeng, L., Hafner, S., Gerard, S., Brune, E., Yadav, R., Shibli, A., Fang, H., Ban, Y., Vergauwen, M., Audebert, N., Nascetti, A.: {PANGAEA}: a global and inclusive benchmark for geospatial foundation models. arXiv:2412.04204  (2025)

\bibitem{nascetti_biomassters_2023}
Nascetti, A., Yadav, R., Brodt, K., Qu, Q., Fan, H., Shendryk, Y., Shah, I., Chung, C.: {BioMassters}: a benchmark dataset for forest biomass estimation using multi-modal satellite time-series. In: NeurIPS Datasets and Benchmarks (2023)

\bibitem{okyay_airborne_2019}
Okyay, U., Telling, J., Glennie, C.L., Dietrich, W.E.: Airborne lidar change detection: an overview of {Earth} sciences applications. Earth-Science Reviews  (2019)

\bibitem{passalacqua_analyzing_2015}
Passalacqua, P., Belmont, P., Staley, D.M., Simley, J.D., Arrowsmith, J.R., Bode, C.A., Crosby, C., DeLong, S.B., Glenn, N.F., Kelly, S.A., Lague, D., Sangireddy, H., Schaffrath, K., Tarboton, D.G., Wasklewicz, T., Wheaton, J.M.: Analyzing high resolution topography for advancing the understanding of mass and energy transfer through landscapes: a review. Earth-Science Reviews  (2015)

\bibitem{pauls_echosat_2026}
Pauls, J., Schrödter, K., Ligensa, S., Schwartz, M., Turan, B., Zimmer, M., Saatchi, S., Pokutta, S., Ciais, P., Gieseke, F.: {ECHOSAT}: estimating canopy height over space and time. arXiv:2602.21421  (2026)

\bibitem{pauls_capturing_2025}
Pauls, J., Zimmer, M., Turan, B., Saatchi, S., Ciais, P., Pokutta, S., Gieseke, F.: Capturing temporal dynamics in large-scale canopy tree height estimation. In: ICML (2025)

\bibitem{qin_3d_2016}
Qin, R., Tian, J., Reinartz, P.: {3D} change detection: approaches and applications. ISPRS Journal of Photogrammetry and Remote Sensing  (2016)

\bibitem{reed_scale-mae_2023}
Reed, C.J., Gupta, R., Li, S., Brockman, S., Funk, C., Clipp, B., Keutzer, K., Candido, S., Uyttendaele, M., Darrell, T.: Scale-{MAE}: a scale-aware masked autoencoder for multiscale geospatial representation learning. In: ICCV (2023)

\bibitem{rolnick_deep_2018}
Rolnick, D., Veit, A., Belongie, S., Shavit, N.: Deep learning is robust to massive label noise. arXiv:1705.10694  (2018)

\bibitem{ronneberger_u-net_2015}
Ronneberger, O., Fischer, P., Brox, T.: U-{Net}: convolutional networks for biomedical image segmentation. In: MICCAI (2015)

\bibitem{roussel_removing_2017}
Roussel, J.R., Caspersen, J., Béland, M., Thomas, S., Achim, A.: Removing bias from {LiDAR}-based estimates of canopy height: accounting for the effects of pulse density and footprint size. Remote Sensing of Environment  (2017)

\bibitem{sai_learning_2009}
Sai, Y., Jinxia, R., Zhongxia, L.: Learning of neural networks based on weighted mean squares error function. In: International Symposium on Computational Intelligence and Design (2009)

\bibitem{sampath_geometric_2016}
Sampath, A., Heidemann, H.K., Stensaas, G.L.: Geometric quality assessment of {LIDAR} data based on swath overlap. ISPRS Archives  (2016)

\bibitem{schwartz_retrieving_2025}
Schwartz, M., Ciais, P., Sean, E., de~Truchis, A., Vega, C., Besic, N., Fayad, I., Wigneron, J.P., Brood, S., Pelissier-Tanon, A., Pauls, J., Belouze, G., Xu, Y.: Retrieving yearly forest growth from satellite data: a deep learning based approach. Remote Sensing of Environment  (2025)

\bibitem{sibona_direct_2017}
Sibona, E., Vitali, A., Meloni, F., Caffo, L., Dotta, A., Lingua, E., Motta, R., Garbarino, M.: Direct measurement of tree height provides different results on the assessment of {LiDAR} accuracy. Forests  (2017)

\bibitem{simeoni_dinov3_2025}
Siméoni, O.: {DINOv3}. arXiv:2508.10104  (2025)

\bibitem{song_learning_2022}
Song, H., Kim, M., Park, D., Shin, Y., Lee, J.G.: Learning from noisy labels with deep neural networks: a survey. IEEE Transactions on Neural Networks and Learning Systems  (2023)

\bibitem{sumbul_bigearthnet_2019}
Sumbul, G., Charfuelan, M., Demir, B., Markl, V.: {BigEarthNet}: a large-scale benchmark archive for remote sensing image understanding. In: IGARSS (2019)

\bibitem{szwarcman_prithvi-eo-20_2025}
Szwarcman, D., Roy, S., Fraccaro, P., G{\'\i}slason, {\TH}.E., Blumenstiel, B., Ghosal, R., Oliveira, P.H.d., Almeida, J.L.d.S., Sedona, R., Kang, Y., Chakraborty, S., Wang, S., Gomes, C., Kumar, A., Truong, M., Godwin, D., Lee, H., Hsu, C.Y., Asanjan, A.A., Mujeci, B., Shidham, D., Keenan, T., Arevalo, P., Li, W., Alemohammad, H., Olofsson, P., Hain, C., Kennedy, R., Zadrozny, B., Bell, D., Cavallaro, G., Watson, C., Maskey, M., Ramachandran, R., Moreno, J.B.: Prithvi-{EO}-2.0: a versatile multi-temporal foundation model for {Earth} observation applications. arXiv:2412.02732  (2025)

\bibitem{tan_triples_2025}
Tan, X., Chen, G., Zhang, X., Wang, T., Wang, J., Wang, K., Miao, T.: {TripleS}: mitigating multi-task learning conflicts for semantic change detection in high-resolution remote sensing imagery. ISPRS Journal of Photogrammetry and Remote Sensing  (2025)

\bibitem{tolan_sub-meter_2023}
Tolan, J., Yang, H.I., Nosarzewski, B., Couairon, G., Vo, H., Brandt, J., Spore, J., Majumdar, S., Haziza, D., Vamaraju, J., Moutakanni, T., Bojanowski, P., Johns, T., White, B., Tiecke, T., Couprie, C.: Very high resolution canopy height maps from {RGB} imagery using self-supervised vision transformer and convolutional decoder trained on aerial lidar. Remote Sensing of Environment  (2024)

\bibitem{tolan_very_2024}
Tolan, J., Yang, H.I., Nosarzewski, B., Couairon, G., Vo, H.V., Brandt, J., Spore, J., Majumdar, S., Haziza, D., Vamaraju, J., Moutakanni, T., Bojanowski, P., Johns, T., White, B., Tiecke, T., Couprie, C.: Very high resolution canopy height maps from {RGB} imagery using self-supervised vision transformer and convolutional decoder trained on aerial lidar. Remote Sensing of Environment  (2024)

\bibitem{tuzlupinar_introducing_2025}
Tuzlupinar, B., Ozelbas, E., Amasyali, M.F., Karaca, A.C.: Introducing {MOSAIC-SEN2-CC}: a multispectral dataset and adaptation framework for remote sensing change captioning. IEEE Journal of Selected Topics in Applied Earth Observations and Remote Sensing  (2025)

\bibitem{wang_disasterm3_2025}
Wang, J., Xuan, W., Qi, H., Liu, Z., Liu, K., Wu, Y., Chen, H., Song, J., Xia, J., Zheng, Z., Yokoya, N.: Disasterm3: A remote sensing vision-language dataset for disaster damage assessment and response  (2025)

\bibitem{wang_constructing_2025}
Wang, Z., Wu, C., Zhang, F., Xia, J.: Constructing an extensible building damage dataset via semi-supervised fine-tuning across 12 natural disasters. Journal of Remote Sensing  (2025)

\bibitem{wing_applying_2010}
Wing, M.G., Eklund, A., Sessions, J.: Applying {LiDAR} technology for tree measurements in burned landscapes. International Journal of Wildland Fire  (2010)

\bibitem{xiao_unified_2018}
Xiao, T., Liu, Y., Zhou, B., Jiang, Y., Sun, J.: Unified perceptual parsing for scene understanding. In: ECCV (2018)

\bibitem{xiong_neural_2025}
Xiong, Z., Wang, Y., Zhang, F., Stewart, A.J., Hanna, J., Borth, D., Papoutsis, I., Saux, B.L., Camps-Valls, G., Zhu, X.X.: Neural plasticity-inspired multimodal foundation model for {Earth} observation. arXiv:2403.15356  (2025)

\bibitem{xu_specialized_2025}
Xu, Z., Gupta, R., Cheng, W., Shen, A., Shen, J., Talwalkar, A., Khodak, M.: Specialized foundation models struggle to beat supervised baselines. In: ICLR (2025)

\bibitem{yang_survey_2025}
Yang, L., Chen, N., Yue, J., Liu, Y., Ma, J., Ghamisi, P., Plaza, A., Fang, L.: Survey of multimodal geospatial foundation models: techniques, applications, and challenges. arXiv:2510.22964  (2025)

\bibitem{yu_evaluating_2024}
Yu, Z., Qi, J., Zhao, X., Huang, H.: Evaluating the reliability of bi-temporal canopy height model generated from airborne laser scanning for monitoring forest growth in boreal forest region. International Journal of Digital Earth  (2024)

\bibitem{zhang_understanding_2017}
Zhang, C., Bengio, S., Hardt, M., Recht, B., Vinyals, O.: Understanding deep learning requires rethinking generalization. In: ICLR (2017)

\bibitem{zhang_hyper-neighborhood_2025}
Zhang, H., Yang, S., Ning, X., He, Y., Huang, X., Zhang, R., Hao, M.: Hyper-neighborhood context-aware transformer network for high-resolution remote sensing change detection. International Journal of Applied Earth Observation and Geoinformation  (2025)

\bibitem{zhang_cross-domain_2023}
Zhang, X., Yu, W., Pun, M.O., Shi, W.: Cross-domain landslide mapping from large-scale remote sensing images using prototype-guided domain-aware progressive representation learning. ISPRS Journal of Photogrammetry and Remote Sensing  (2023)

\bibitem{zhao_domain-adaptive_2021}
Zhao, A., Ding, M., Lu, Z., Xiang, T., Niu, Y., Guan, J., Wen, J.R.: Domain-adaptive few-shot learning. In: WACV (2021)

\bibitem{zhao_utility_2018}
Zhao, K., Suarez, J.C., Garcia, M., Hu, T., Wang, C., Londo, A.: Utility of multitemporal lidar for forest and carbon monitoring: tree growth, biomass dynamics, and carbon flux. Remote Sensing of Environment  (2018)

\bibitem{zhou_segchange-r1_2025}
Zhou, F.: {SegChange-R1}: {LLM}-augmented remote sensing change detection. arXiv:2506.17944  (2025)

\end{thebibliography}

\end{document}